%% file: main.tex
\documentclass[sigconf]{acmart}

\def\BibTeX{{\rm B\kern-.05em{\sc i\kern-.025em b}\kern-.08emT\kern-.1667em\lower.7ex\hbox{E}\kern-.125emX}}

\usepackage{amsmath}
\usepackage{algorithm}
\usepackage{algorithmic}
    
\copyrightyear{2019} 
\acmYear{2019} 
\setcopyright{acmlicensed}
\acmConference[KDD '19]{The 25th ACM SIGKDD Conference on Knowledge Discovery and Data Mining}{August 4--8, 2019}{Anchorage, AK, USA}
\acmBooktitle{The 25th ACM SIGKDD Conference on Knowledge Discovery and Data Mining (KDD '19), August 4--8, 2019, Anchorage, AK, USA}
\acmPrice{15.00}
\acmDOI{10.1145/3292500.3330739}
\acmISBN{978-1-4503-6201-6/19/08}
\begin{document}

%
\title{Learning a Unified Embedding \\ for Visual Search at Pinterest}

%

\author{Andrew Zhai${^{^{1}}}$, Hao-Yu Wu${^{^{1}}}$, Eric Tzeng${^{^{12}}}$, Dong Huk Park${^{^{12}}}$, Charles Rosenberg${^{^{1}}}$}
\affiliation{${^{^1}}$ Visual Discovery, Pinterest \qquad ${^{^2}}$University of California, Berkeley}
\email{{andrew,rexwu,etzeng,dhukpark,crosenberg}@pinterest.com}

%

\renewcommand{\shortauthors}{Zhai et al.}
\newcommand{\seth}[1]{\textcolor{red}{Seth: #1}}

%

\include{main_abstract}

%
%


%

%
\begin{teaserfigure}
\begin{center}
\includegraphics[width=1.0\linewidth]{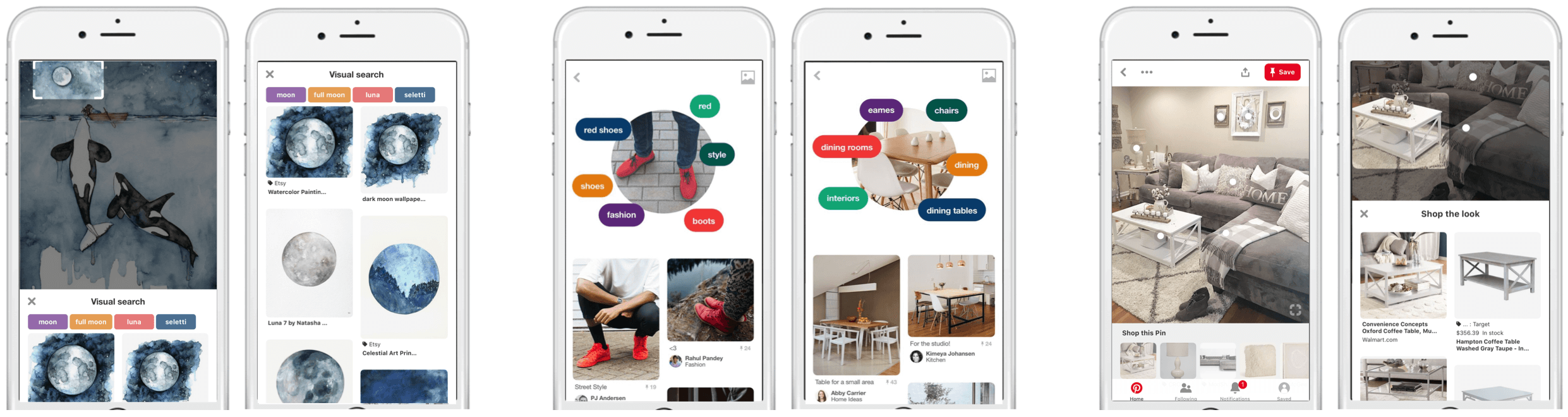}
\end{center}
   \caption{Three visual search products on Pinterest (Flashlight, Lens, and Shop-the-Look) allowing users to browse content related to web or camera images and search for exact products within home scenes for shopping. Visual search is one of the fastest growing products at Pinterest with over 600M searches per month.}
\label{fig:applications}
\end{teaserfigure}

%
\maketitle

{\fontsize{8pt}{8pt} \selectfont
\textbf{ACM Reference Format:} \\
Andrew Zhai, Hao-Yu Wu, Eric Tzeng, Dong Huk Park, Charles Rosenberg. 2019. Learning a Unified Embedding for Visual Search at Pinterest. In \textit{The 25th ACM SIGKDD Conference on Knowledge Discovery and Data Mining (KDD '19), August 4--8, 2019, Anchorage, AK, USA.} ACM, New York, NY, USA, 9 pages. https://doi.org/10.1145/3292500.3330739  }

\input{main_introduction.tex}

\input{main_related.tex}

\input{main_method.tex}
\input{main_experiments.tex}
\input{main_conclusion.tex}

%

\bibliographystyle{ACM-Reference-Format}
\bibliography{main}

%

\end{document}

%% file: main_abstract.tex
\begin{abstract}
At Pinterest, we utilize image embeddings throughout our search and recommendation systems to help our users navigate through visual content by powering experiences like browsing of related content and searching for exact products for shopping. In this work we describe a multi-task deep metric learning system to learn a single unified image embedding which can be used to power our multiple visual search products. The solution we present not only allows us to train for multiple application objectives in a single deep neural network architecture, but takes advantage of correlated information in the combination of all training data from each application to generate a unified embedding that outperforms all specialized embeddings previously deployed for each product. We discuss the challenges of handling images from different domains such as camera photos, high quality web images, and clean product catalog images. We also detail how to jointly train for multiple product objectives and how to leverage both engagement data and human labeled data.  In addition, our trained embeddings can also be binarized for efficient storage and retrieval without compromising precision and recall. Through comprehensive evaluations on offline metrics, user studies, and online A/B experiments, we demonstrate that our proposed unified embedding improves both relevance and engagement of our visual search products for both browsing and searching purposes when compared to existing specialized embeddings.  Finally, the deployment of the unified embedding at Pinterest has drastically reduced the operation and engineering cost of maintaining multiple embeddings while improving quality.
\end{abstract}

%% file: main_introduction.tex
\section{Introduction}
 Following the explosive growth in engagement with online photography and videos, visual embeddings have become increasingly critical in search and recommendation systems. Content based image retrieval systems (visual search) is one prominent application that heavily relies on visual embeddings for both ranking and retrieval as users search by providing an image. In recent years, visual search has proliferated across a portfolio of companies including Alibaba's Pailitao~\cite{visualsearchalibaba}, Pinterest Flashlight and Lens~\cite{visualdiscoverypinterest}~\cite{jing15}. Google Lens, Microsoft's Visual Search~\cite{visualsearchbing}, and Ebay's Visual Shopping~\cite{visualsearchebay}. These applications support a wide spectrum of use cases from Shopping where a user is \textit{searching} for the exact item to Discovery~\cite{visualdiscoverypinterest} where a user is \textit{browsing} for inspirational and related content. These interactions span both real world (phone camera) and online (web) scenarios. 


Over 250M users come to Pinterest monthly to discover ideas for recipes, fashion, travel, home decor, and more from our content corpus of billions of Pins. To facilitate discovery, Pinterest offers a variety of products including text-to-pin search, pin-to-pin recommendations~\cite{relatedpins}, and user-to-pin recommendations. Throughout the years, we've  built a variety of visual search products (Figure~\ref{fig:applications}) including Flashlight (2016), Lens (2017), and automated Shop-the-Look (2018) to further empower our users to use images (web or camera) as queries for general browsing or shopping~\cite{visualdiscoverypinterest}. With over 600 million visual searches per month and growing, visual search is one of the fastest growing products at Pinterest and of increasing importance.

We have faced many challenges training and deploying generations of visual embeddings over the years throughout the search and recommendation stack at Pinterest. The difficulties can be summarized to the following four aspects:

\textbf{Different Applications have Different Objectives}: Pinterest uses image embeddings for a variety of tasks including retrieval (pin and image), ranking (text, pin, user, image queries), classification or regression (e.g. neardup classification, click-through-rate prediction, image type), and upstream multi-modal embedding models (PinSAGE~\cite{pinsage}). One observation we made with these multiple applications is that optimization objectives are not the same. Take our visual search products in Figure~\ref{fig:applications} as examples: Flashlight optimizes for browsing relevance \textit{within} Pinterest catalog images. Lens optimizes for browsing Pinterest catalog images from camera photos; hence overcoming the domain shift of camera to Pinterest images is necessary. Finally Shop-the-Look optimizes for searching for the \textit{exact} product from objects in a scene for shopping.

\textbf{Embedding Maintenance/Cost/Deprecation}: Specialized visual embeddings per application are the clearest solution to optimizing for multiple consumers and is the paradigm operated at Pinterest prior to 2018. This however has significant drawbacks. For our visual search products alone, we developed three specialized embeddings. 
As image recognition architectures are evolving quickly~\cite{alexnet12}~\cite{vggnet2014}~\cite{kaiming16}~\cite{resnext}~\cite{senet}, we want to iterate our three specialized embeddings with modern architectures to improve our three visual search products. Each improvement in embedding requires a full back-fill for deployment, which can be prohibitively expensive. 
In practice, this situation is further exacerbated by downstream dependencies (e.g. usage in pin-to-pin ranking~\cite{relatedpins}) on various \textbf{specific} versions of our embeddings, leading us to incrementally continue to extract multiple generations of the \textit{same} specialized embeddings. All these considerations make the unification of specialized embeddings into one general embedding very attractive, allowing us clear tracking of external dependency in one lineage along with scalability to support future optimization targets.

\textbf{Effective Usage of Datasets}: At Pinterest, there are various image data sources including engagement feedback (e.g. pin-to-pin click data~\cite{relatedpins}, Pin-to-Board graph when users save Pins into collections called Boards~\cite{pixie}) and human curation. When training specialized embeddings for a specific task, deciding what datasets to use or collect is a non-trivial issue, and the choice of data source is often based on human judgement and heuristics which could be suboptimal and not scalable. Multi-task learning simplifies this by allowing the model to learn what data is important for which task through end-to-end training. Through multi-task learning, we want to minimize the amount of costly human curation while leveraging as much engagement data as possible. 

\textbf{Scalable and Efficient Representation}: With billions of images and over 250M+ monthly active users, Pinterest has a requirement for an image representation that is cheap to store and also computationally efficient for common operations such as distance for image similarity. To leverage the large amount of training data that we receive from our user feedback cycles, we also need to build efficient model training procedures. As such, scalablity and efficiency are required both for inference and training.

In our paper, we describe our implementation, experimentation, and productionization of a unified visual embedding, replacing the specialized visual embeddings at Pinterest. The main contributions of this paper are (1) we present a scalable multi-task metric learning framework (2) we present insights into creating efficient multi-task embeddings that leverage a combination of human curated and engagement datasets (3) we present lessons learned when scaling training of our metric learning method and (4) comprehensive user studies and AB experiments on how our unified embeddings compare against the existing specialized embeddings across all visual search products in Figure~\ref{fig:applications}.

%% file: main_related.tex
\section{Related Works}
\subsection{Visual Search Systems}
Visual search has been adopted throughout industry with Ebay~\cite{visualsearchebay}, Microsoft~\cite{visualsearchbing}, Alibaba~\cite{visualsearchalibaba}, Google (Lens), and Amazon launching their own products. There has also been an increasing amount of research on domain-specific image retrieval systems such as fashion~\cite{fashion_rec} and product~\cite{bell15productnet} recommendations. Compared to others, Pinterest has not just one but a variety of visual search products (Figure~\ref{fig:applications}), each with different objectives. We focus on addressing the challenges of unifying visual embeddings across our visual search products.

\subsection{Metric Learning}
\label{subsec:related_metric}
Standard metric learning approaches aim to learn image representations through the relationships between images in the form of \textit{pairs}~\cite{siamese}~\cite{bell15productnet} or \textit{triplets}~\cite{triplet}~\cite{facenet}. Similarity style supervision are used to train the representation such that similar images are close in embedding space and dissimilar images apart. Sampling informative negatives is an important challenge of these pair or triplet based approaches, a focus of recent methods such as ~\cite{songCVPR16}~\cite{npairNIPS2016}~\cite{samplingmatters}. 

An alternative approach to metric learning are classification based methods~\cite{nofusslearning}\cite{andrewhao} which alleviate the need of negative sampling. These methods have recently have been shown to achieve SOTA results across a suite of retrieval tasks~\cite{andrewhao} compared with pair or triplet based methods. Given the simplicity and effectiveness of formulating metric learning as classification, we build off the architecture proposed in~\cite{andrewhao} and extend it to multi-task for our unified embeddings.

\subsection{Multi-Task Learning}
Multi-task learning aims to learn one model that provides multiple outputs from one input~\cite{ubermultitask}~\cite{multitasksynthetic}. By consolidating multiple single-task models into one multi-task model, previous work have seen both efficiency~\cite{ubermultitask} and performance~\cite{multitaskloss}~\cite{crossstitch}~\cite{multitasksynthetic} improvements on each task due to inherent complementary structure that exists in separate visual tasks~\cite{taskonomy}. Prior work also investigate how to learn to balance multiple loss objectives to optimize performance~\cite{multitaskloss}~\cite{gradnorm}. In the context of metric learning, \cite{csn} and \cite{Zhao2018AMM} explore the idea of learning a conditional mask for each task to modulate either the learned embedding or the internal activations. In our paper, we experiment with multi-task metric learning and evaluate its effects on our web scale visual search products. 

%% file: main_method.tex
\section{Method}

\subsection{Problem Setup}

Pinterest is a visual discovery platform in which the contents are predominantly images. To empower the users on Pinterest to \textit{browse} visually inspired contents and \textit{search} for an exact item in the image for shopping, we have built three services shown in Figure \ref{fig:applications}: Flashlight, Lens, and Shop-The-Look (STL). Flashlight enables the users to start from the images on Pinterest (or web), and recommends relevant Pins inspired by the input images for the users to browse. Similarly, Lens aims to recommend visually relevant Pins based on the photos our users take with their cameras. STL, on the other hand, searches for products which are best match to the input images for the users to shop. The three services either serve images from different domains (web images v.s. camera photos), or with different objectives (browsing v.s. searching).

\begin{figure}[t]
\begin{center}
\includegraphics[width=1.0\linewidth]{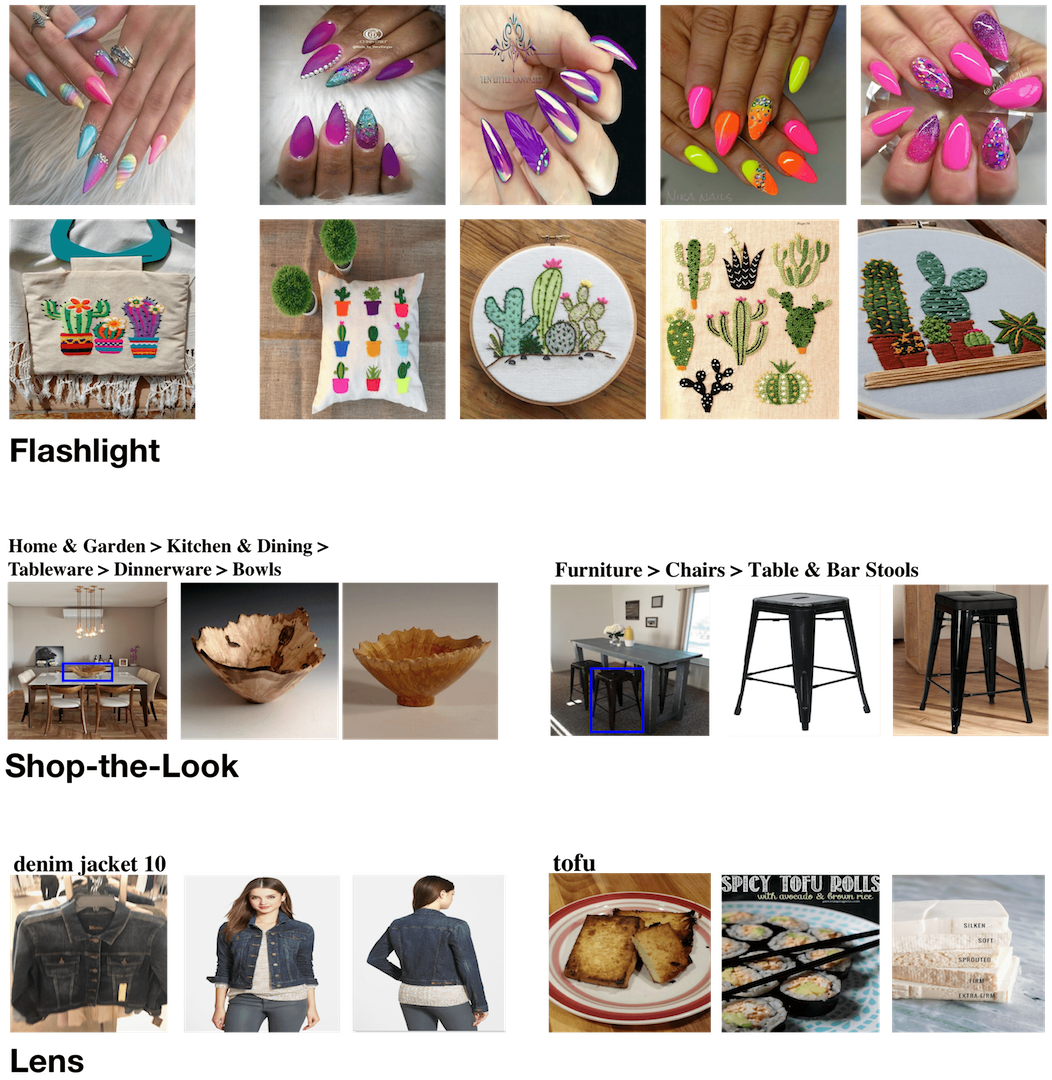}
\end{center}
   \caption{Visualization of our training datasets}
\label{fig:training_datasets}
\end{figure}

With both cost and engineering resource constraints and the interest of improved performance, we aim to learn one unified image embedding that can perform well for all three tasks. In essence, we would like to learn high quality embeddings of images that can be used for both browsing and searching recommendations. The relevance or similarity of a pair of images is represented as the distance between the respective embeddings. In order to train such embeddings, we collected a dataset for each task addressing its specific objective (Figure~\ref{fig:training_datasets}), and frame the problem as multi-task metric learning that jointly optimizes the relevance for both browsing and searching. We will describe how we collect the dataset for each task, the detailed model architecture, and how we set up multi-task training in the following sections.

\subsection{Training Data}
\label{subsec:dataset}

We describe our training datasets and show some examples in Figure~\ref{fig:training_datasets}.

\subsubsection{Flashlight Dataset}
The Flashlight dataset is collected to define browse relevance for Pinterest (web) images. As a surrogate for relevance, we rely on engagement through feedback from our users in a similar manner to Memboost described in~\cite{relatedpins} to generate the dataset. Image sets are collected where a given query image has a set of related images ranked via engagement, and we assign each image set a label (unique identifier) that is conceptually the same as semantic class label. We apply a set of strict heuristics (e.g. number of impressions and interactions for each image in the set) to reduce the label noise of the dataset, resulting in around 800K images in 15K semantic classes.

\subsubsection{Lens Dataset}
The Lens dataset is collected to define browse relevance between camera photo images and Pinterest images. When collecting the training dataset for prototyping Lens, we found that camera photo engagement on Pinterest is very sparse and as such any dataset collected via user engagement would be too noisy for training. The main obstacle we need to overcome is the domain shift between camera photos and Pinterest images, so for the training set, we collected a human labeled dataset containing 540K images with 2K semantic classes. These semantic classes range from broad categories (e.g. tofu) to fine-grained classes (e.g. the same denim jacket in camera and product shots). Most importantly, this dataset contains a mix of product, camera, and Pinterest images under the same semantic label so that the embeddings can learn to overcome the domain shifts. 

\subsubsection{Shop-The-Look Dataset}
\label{data:shopping}
The Shop-The-Look dataset is collected to define search relevance for an object in a home decor scene to its product match. To bootstrap the product, we collected a human labeled dataset containing 340K images with 189 product class label (e.g. Bar Stools) and 50K instance labels. Images with the same instance label are either exact matches or are very similar visually as defined by an internal training guide for criteria such as design, color, and material.

\subsection{Model Architecture}
\label{sec:model_architecture}
\begin{figure*}[t]
\begin{center}
\includegraphics[width=1.0\linewidth]{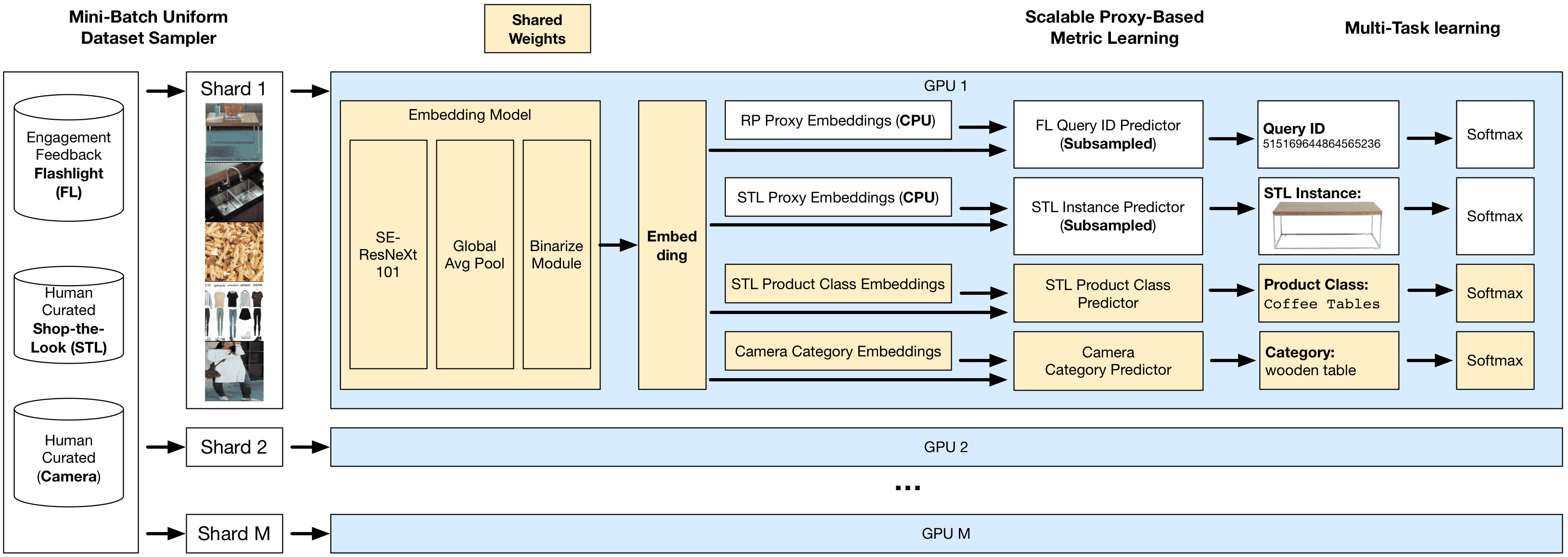}
\end{center}
   \caption{The overall architecture for multi-task metric learning network. The proposed classification network as proxy-based metric learning is simple and flexible for multi-task learning. Our proposed method also has the binarization module to make learned embedding memory efficient, and the subsampling module is scalable to support large number of classes.}
\label{fig:architecture}
\end{figure*}

Figure~\ref{fig:architecture} illustrates our overall setup for multi-task metric learning architecture. We extend the classification-based metric learning approach of Zhai et al. \cite{andrewhao} to multi-task. All the tasks share a common base network until the embedding is generated, where each task then splits into their own respective branches. Each task branch is simply a fully connected layer (where the weights are the proxies) followed by a softmax (no bias) cross entropy loss. There are four softmax tasks, Flashlight class, Shop-the-Look (STL) product class, STL instance class, and Lens category class. For the Flashlight class and STL instance tasks, the proxies are subsampled using our subsampling module before input to the fully connected layer for efficient training.

There are two essential modules for web-scale applications: a subsampling module to make our method \textit{scalable} to hundreds of thousands of classes, and binarization module to make learned embedding storage and operation \textit{efficient}.



\subsubsection{Subsampling Module}
\label{subsubsec:subsampling}
\begin{figure}[t]
\begin{center}
\includegraphics[width=1.0\linewidth]{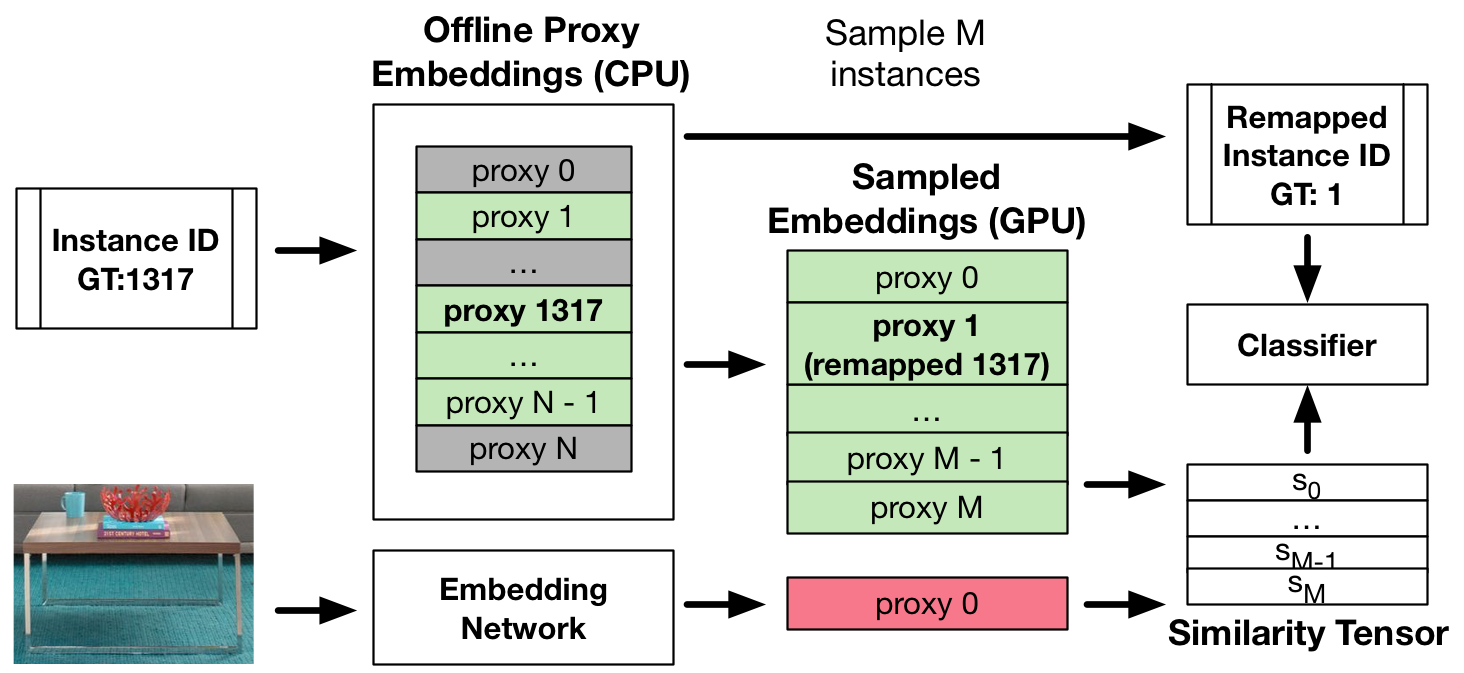}
\end{center}
   \caption{Visualization of the subsampling module.}
\label{fig:sampling}
\vspace{-4mm}
\end{figure}

Given N images in a batch and M proxies to target each with an embedding dimension of D, to compute the similarity (dot product) of embeddings to proxies, we do a NxD by DxM matrix multiplication in the fully connected layer. As such, computation increases with the number of proxy classes (M), an undesirable property as we scale M. Furthermore, we may not even be able to fit the proxy bank (MxD matrix) in GPU memory as we scale M to millions of classes (user engagement training data can easily generate this many classes). To address these issues, we store the proxy bank in CPU RAM (more available than GPU memory and disk can be used later if necessary) along with also implementing class \textbf{subsampling}. As shown in Figure~\ref{fig:sampling},  for each training batch, the subsampling module samples a subset of all classes for optimization. The sampled subset is guaranteed to have all the ground truth class labels of the images in the training batch (the label index slot will change however to ensure the index is within bounds of the number of classes sampled). The pseudocode for the forward pass is provided in Algorithm~\ref{alg:subsampling}. During the training forward pass for efficiency, the proxies of the sampled classes are moved to GPU for computation asynchronously while the embedding is computed from the base network. The softmax loss only considers the sampled classes. For example, if we subsample only 2048 classes for each iteration, the maximum loss from random guessing is $ln(2048) \approx 7.62$.

\begin{algorithm}
\caption{Subsampling Proxy Indices}\label{alg:subsampling}
\begin{flushleft}
\textbf{Input:} targets, num\_samples\\
\textbf{Output:} sampled\_proxy\_idx, remapped\_targets\\ 
\end{flushleft}
\begin{algorithmic}[1]
\REQUIRE $len(targets) \leq num\_samples$
\STATE $sampled\_proxy\_idx \leftarrow set(targets)$
\WHILE{$len(sampled\_proxy\_idx) \leq num\_samples$} 
    
        \STATE{$s \leftarrow sample(all\_labels)$}
        \IF{$s \notin sampled\_proxy\_idx$} 
            \STATE {$sampled\_proxy\_idx.add(s)$}
        \ENDIF
\ENDWHILE

\STATE $sampled\_proxy\_idx \leftarrow list(sampled\_proxy\_idx)$
\STATE $remapped\_targets \leftarrow list([])$
\FORALL{$t \in targets$}
    \FORALL{$index, label \in enumerate(sampled\_proxy\_idx)$}
        \IF{$t = label$} 
            \STATE {$remapped\_targets.add(index)$}
        \ENDIF
    \ENDFOR
\ENDFOR

\RETURN $sampled\_proxy\_idx, remapped\_targets$
\end{algorithmic}
\end{algorithm}

\subsubsection{Binarization Module}
\label{sec:motiv_binary}
 At Pinterest, we have a growing corpus of billions of images and as such we need efficient representations to (1) decrease cold storage costs (e.g. cost of AWS S3) (2) reduce bandwidth for downstream consumers (e.g. I/O costs to fully ingest the embeddings into map reduce jobs) (3) improve the latency of real-time scoring (e.g. computing the similarity of two embeddings for ranking). In the prior work~\cite{andrewhao}, we see that embeddings learned from the classification based metric learning approach can be binarized by thresholding at zero with little drop in performance.
 
We consider everything after the global pooling layer but before the task specific branches as the "binarization" module. In this work, instead of LayerNorm as proposed in ~\cite{andrewhao}, we propose to use GroupNorm~\cite{groupnorm} that is better suited for multi-task applications. The empirical results are provided in Section~\ref{sec:binarization}.

\subsection{Model Training}
Shown in Figure~\ref{fig:architecture}, we train our model in a multi-tower distributed setup. We share parameters throughout the network with the exception of the sparse proxy parameters. Each node (out of eight) has its own full copy of the CPU Embedding bank as sparse parameters cannot be distributed at this moment by the PyTorch framework. Empirically, not distributing the sparse parameters led to no performance impact.

\subsubsection{Mini-Batch and Loss for Multi-task}
\label{sec:multitask_arch}
For every mini-batch, we balance a uniform mix of each of the datasets with an epoch defined by the iterations to iterate through the largest dataset. Each dataset has its own indepedent tasks so we ignore the gradient contributions of images on tasks that it does not have data for. The losses from all the tasks are assigned equal weights and are summed for backward propagation.

\subsubsection{Sparse Tensor Optimization}
We represent the proxy banks that are sparsely subsampled as sparse tensors. This avoids expensive dense gradient calculation for all the proxies during the backward propagation.

An additional optimization is the handling of momentum. By enabling momentum update on the sparse tensor, the sparse gradient tensors will be aggregated and become expensive dense gradient updates. Since momentum is crucial for deep neural network optimization, we approximate the momentum update for sparse tensor by increasing the learning rate. Assuming we choose $momentum = 0.9$, the gradients of the current iteration $\hat{G}$ will roughly have net update effect of $10x$ learning rate over the coarse of training: $$\sum_{n=1}^{\infty} lr\times\hat{G}\times0.9^{n} = 10.0\times lr \times \hat{G}$$ Although increasing learning rate will affect the optimization trajectory on the loss function surface, thus affect the subsequent gradients, we find this approximation decreases our training time by 40\% while retaining comparable performance.

\subsection{Model Deployment}
We train our models using the PyTorch framework and deploy models through PyTorch to ONNX to Caffe2 conversion. For operations where ONNX does not have a compatible representation for, we directly use the ATen operator, a shared backend between PyTorch and Caffe2, to bypass and deploy our Caffe2 model.




%% file: main_experiments.tex
\section{Experiments}
In this section, we measure the performance of our method on a suite of evaluations including offline measurement, human judgements, and A/B experiments. We demonstrate the efficacy and the impact of our unified embeddings at Pinterest.

\subsection{Implementation}
\label{exp:implementation}
Our model is trained using PyTorch on one p3.16xlarge Amazon EC2 instances with eight Tesla V100 graphic cards. We use the DistributedDataParallel implementation provided by the PyTorch framework for distributed training.

We train our models with largely the same hyperparameters as~\cite{andrewhao} with SE-ResNeXt101~\cite{senet} as the base model pre-trained on ImageNet ILSVRC-2012~\cite{imagenet_cvpr09}. We use SGD with momentum of 0.9, weight decay of 1e-4, and gamma of 0.1. We start with a base learning rate of 0.08 (0.01 x 8 from the linear scaling heuristic of~\cite{imagenet1hr}) and train our model for 1 epoch by updating only new parameters for better initialization with a batch size of 128 per GPU. We then train end-to-end with a batch size of 64 per GPU and apply the gamma to reduce learning rate every 3 epochs for a total of 9 epochs of training (not counting initialization). During training, we apply horizontal mirroring, random crops, and color jitter from resized 256x256 images while during testing we center crop to a 224x224 image from the resized image.

\input{main_offline.tex}

\input{main_pinterest_exp.tex}

%% file: main_offline.tex
\subsection{Offline Evaluation}
Offline measurements are the first consideration when iterating on new models. For each product of interest (Shop-the-Look, Flashlight, and Lens), we have a retrieval evaluation dataset. Some are derived based on the training data while others are sampled according to usage in product. The difference in approach is due to either boostrapping a new product versus improving an existing one.

The evaluation dataset for Shop-the-Look is generated through human curation as we looked to build this new product in 2018. We sampled home decor scenes according to popularity (\# of closeups) on Pinterest and used human curation to label bounding boxes of objects in scene along with ground truth whole image product matches to the objects (criteria determined by our developed internal training guide). This resulted in 600 objects with 1421 ground truth product matches. We measure Precision@1~\cite{evaluation_metrics} for evaluation where for each of the objects, we extract its embedding and generate the top nearest neighbor result in a corpus of 51421 product images (50K sampled from our product corpus + the ground truth whole product images). Precison@1 is then the percent of objects that have a ground truth whole product image retrieved.

The evaluation datasets for Flashlight and Lens are generated through user engagement. For Flashlight, we sampled a random class subset of the Flashlight training data (Section~\ref{subsec:dataset}), and randomly divided it into 807 images for queries and 42881 images for the corpus across classes. For Lens, we generated the evaluation dataset using user engagement in the same manner as the Flashlight training dataset (Section~\ref{subsec:dataset}) but filtering the query images to be camera images with human judgement. This resulted in 1K query images and 49K images for the corpus across classes. As these datasets are generated from noisy user engagement, we use the evaluation metric of Average Precision@20 where we take the average of Precision@1 to Precision@20 (Precision@K as defined in~\cite{visualdiscoverypinterest}). Empirically we have found that though these evaluations are noisy, significant improvements in this Average P@20 have correlated with improvements in our online A/B experiment metrics.

\subsubsection{Binarization}
\label{sec:binarization}
We experiment with model architectures in Table~\ref{tab:binary}. We are primarily interested in \textbf{binarized} embedding performance as described in Section~\ref{sec:motiv_binary}. An alternative to binary features for scalability is to learn low dimensional float embeddings. Based on our prior work~\cite{andrewhao} however, we found that for the same storage cost, learning binary embedding led to better performance than learning float embeddings. 

Our baseline approach in Table~\ref{tab:binary} was to apply the LayerNorm (ln) with temperature of 0.05 and NormSoftmax approach of~\cite{andrewhao} with a SE-ResNeXt101 featurizer and multi-task heads (Section~\ref{sec:model_architecture}). We noticed a significant drop in performance between raw float and binary features. We experimented with variations of the architecture including: Softmax (sm) to remove L2 normalized embedding constraint, GroupNorm~\cite{groupnorm} (group=256) for more granularity in normalization, ReLU (r) to ignore negative magnitudes, and Dropout (p=0.5) for learning redundant representations. As shown, our final binarized multi-task embeddings performs favorably to the raw float features baseline.

\begin{table}
\begin{center}
\begin{tabular}{l r r r r r }
\hline
Model & STL & Flashlight & Lens \\
& P@1 & Avg P@20 & Avg P@20  \\
\hline\hline
~\cite{andrewhao} Baseline (f) & 47.5 & 60.1 & 18.6 & \\
\hline
~\cite{andrewhao} Baseline (b) & 41.9 & 55.6 & 17.7 \\
+sm + gn (b) & 48.4 & 59.3 & \underline{17.8} \\
+sm + gn + r (b) & \underline{49.7} & \textbf{61.1} & 17.6 \\
+sm + gn + r + dp (b) (Ours) & \textbf{52.8} & \underline{60.2} & \textbf{18.4} \\
\hline
\end{tabular}
\end{center}
\caption{Model architecture experiments on offline evaluations (f = float, b = binary). We compare binary embeddings for deployment.}
\label{tab:binary}
\vspace{-5mm}
\end{table}

\subsubsection{Multi-Task Architecture Ablations}

\begin{table}
\begin{center}
\begin{tabular}{l r r r r r }
\hline
Model & STL & Flashlight & Lens \\
& P@1 & Avg P@20 & Avg P@20  \\
\hline\hline
Ours & 52.8 & 60.2 & 18.4 \\
+ No Dataset Balancing & 44.4 & 57.8 & 18.6 \\
+ GradNorm & 47.2 & 57.8 & 17.2 \\
\hline
\end{tabular}
\end{center}
\caption{Multi-Task experiments on offline evaluations.}
\label{tab:multitask}
\vspace{-5mm}
\end{table}

We show our multi-task experiment results in Table~\ref{tab:multitask}. Instead of uniform sampling of each dataset in a mini-batch (Section~\ref{sec:multitask_arch}), we experiment with sampling based on dataset size. Instead of assigning equal weights to all task losses, we experiment with GradNorm~\cite{gradnorm} to learn the weighting. We see our simple approach achieved the best balance of performance. 


\subsubsection{Multi-Task Dataset Ablations}
\label{sec:ablation}
We look at how training with multiple datasets affects our unified embeddings. In Table~\ref{tab:ablation}, we compare our multi-task embedding trained with all three datasets against embeddings (using the same architecture) trained with each dataset independently. When training our embedding with one dataset, we ensure that the total iterations of images over the training procedure is the same as when training with all three datasets. We see that multi-task improves all three retrieval metrics compared with the performance of embeddings trained on a single dataset.

\begin{table}
\begin{center}
\begin{tabular}{l r r r r r}
\hline
Dataset & STL & Flashlight & Lens \\
& P@1 & Avg P@20 & Avg P@20 \\
\hline\hline
Shop-the-Look (S) & \underline{49.2} & 42.1 & 14.7  \\
Flashlight (F) & 11.0 & \underline{53.4} & 16.1 \\
Lens (L) &  26.2 & 47.8 & \underline{18.2}  \\
All (S + F + L) & \textbf{52.8} & \textbf{60.2} & \textbf{18.4} \\
\hline
\end{tabular}
\end{center}
\caption{Ablation study on datasets. We train specialized embeddings for each training dataset and compare with our unified embedding trained on all training datasets in multi-task. We compare against binary feature performance.}
\label{tab:ablation}
\vspace{-7mm}
\end{table}

\subsubsection{Unified vs Specialized embeddings}
We compare our unified embedding against the previous specialized embeddings~\cite{visualdiscoverypinterest} deployed in Flashlight, Lens, and Shop-the-Look in Table~\ref{tab:baseline_comparisons}. We also include a SENet~\cite{senet} pretrained on ImageNet baseline for comparison. We see our unified embedding outperforms both the baseline and all specialized embeddings for their respective tasks.

Although the unified embeddings compare favorably to the specialized embeddings, the model architectures of these specialized embeddings are fragmented. Flashlight embedding are generated from a VGG16~\cite{vggnet2014} FC6 layer of 4096 dimensions. Lens embedding are generated from a ResNeXt50~\cite{resnext} final pooling layer of 2048 dimensions. Shop-the-Look embedding are generated from a ResNet101~\cite{kaiming16} final pooling layer of 2048 dimensions. This fragmentation is undesirable as each specialized embedding can benefit from updating the model architecture to our latest version as seen in the dataset ablation studies in Section~\ref{sec:ablation}. However in practice, this fragmentation is the direct result of challenges in \textbf{embedding maintenance} from focusing on different objectives at different times in the past. Beyond the improvements in offline metrics from multitask as seen in the Ablation study, the engineering simplification of iterating only one model architecture is an additional win.

\begin{table}
\begin{center}
\begin{tabular}{l r r r r r}
\hline
Model & STL & Flashlight & Lens \\
& P@1 & Avg P@20 & Avg P@20 \\
\hline\hline
Old Shop-the-Look & 33.0 & - & - \\
Old Flashlight & - & 53.4 & - \\ 
Old Lens & - & - & 17.8 \\
ImageNet & 5.6 & 33.1 & 15.0 \\   
Ours & \textbf{52.8} & \textbf{60.2} & \textbf{18.4} \\
\hline
\end{tabular}
\end{center}
\caption{Our binary unified embedding against the existing specialized binary embeddings for each application. We include an ImageNet baseline using a pre-trained SENet\cite{senet}}
\label{tab:baseline_comparisons}
\vspace{-7mm}
\end{table}

%% file: main_pinterest_exp.tex
\subsection{Human Judgements}
Offline evaluations allow us to measure, on a small corpus, the improvements of our embeddings alone. Practical information retrieval systems however are complex, with retrieval, lightweight score, and ranking components~\cite{jing15} using a plethora of features. As such it is important for us to measure the impact of our embeddings end-to-end in our retrieval systems. To compare our unified embeddings with the productionized specialized embeddings for human judgement and A/B experiments, we built separate clusters for each visual search product where the difference between the new and production clusters are the unified vs specialized embeddings.

\begin{figure}[t]
\begin{center}
\includegraphics[width=0.9\linewidth]{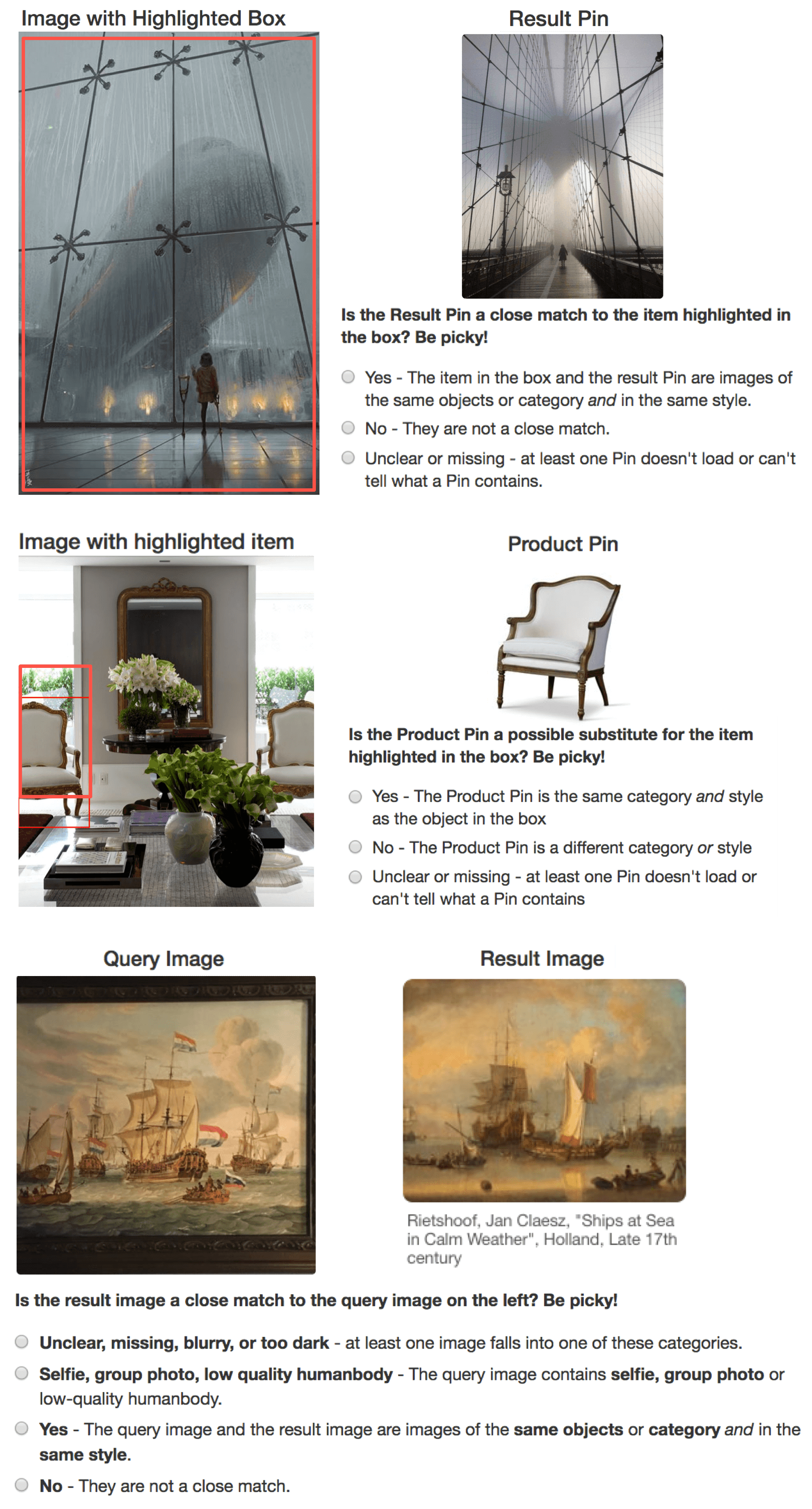}
\end{center}
   \caption{Human Judgement Task templates for Flashlight (top), Shop-The-Look (middle), and Lens (bottom).}
\label{fig:human_eval}
\end{figure}

At Pinterest, we rely on human judgement to measure the relevance of our visual search products and use A/B experiments to measure engagement. For each visual search product, we built relevance templates (Figure~\ref{fig:human_eval}) tuned with an internal training guide for a group of internal workers (similar to Amazon Mechanical Turk) where we describe what \textit{relevance} means in the product at hand with a series of expected (question, answer) pairs. To control quality, for a given relevance template we ensure that workers can achieve 80\%+ precision in their responses against a golden set of (question, answer) pairs. We further replicate each question 5 times, showing 5 different workers the same question and aggregating results with the majority response as the final answer. We also record worker consistency (WC) across different sets of jobs measuring given the same question multiple times, what percent of the questions were answered the same across jobs.

Questions for judgement are generated from a traffic weighted sample of queries for each product. Given each query (Pin image + crop for Flashlight, Camera image for Lens, and Pin image + crop for Shop-the-Look), we send a request to each product's visual search system to generate 5 results per query. Each (query, result) pair forms a question allowing us to measure Precision@5. We generate two sets of 10K (question, answer) tasks per visual search product, one on the existing production system with the specialized embedding and another on the new cluster with our unified embedding. Our human judgement results are shown in Table~\ref{tab:flashlight_lens_eval} for Flashlight and Lens and Table~\ref{tab:stl_eval} for Shop-the-Look. As we can see, our new unified embeddings significantly improve the relevance of all our visual search products.

One hypothesis for these significant gains beyond better model architecture is that combining the three datasets covered the weaknesses of each one independently. Flashlight allows crops as input while the engagement generated training data are whole images. Leveraging the Shop-the-Look dataset with crops helps bridge this domain shift gap of crop to whole images. Similarly for Lens, though the query is a camera image and we need to address the camera to pin image domain shift, the content in the corpus are all Pinterest content. As such additionally using Pinterest corpus training data like Flashlight's can allow the embedding to not only handle camera to pin image matches but also better organize Pinterest content in general. Such dataset interactions are not immediately clear when training specialized embeddings. By learning a single unified embedding with all our training datasets, we let the model training learn how to \textbf{effectively use the datasets} for each task.

\begin{table}
\begin{center}
\begin{tabular}{l | c | c | c | c  | c}
\hline
Application & Win & Lose & Draw & P@5 Delta & WC \\
\hline
Flash. (old vs new) & 41.3\% & 13.9\% & 44.8\% & +22.2\% & 98.0\% \\
Lens (old vs new) & 54.0\% & 7.9\%  & 38.0\% & +110.1\% & 92.9\% \\
\hline
\end{tabular}
\end{center}
\caption{Human Judgements for Flashlight and Lens measuring Precision@5 delta comparing unified embedding vs existing specialized embedding along with the percent of queries that are better (Win), worse (Lose), or have the same (Draw) Precision@5. We see that our new unified embedding significantly improves the relevance of the two products.}
\label{tab:flashlight_lens_eval}
\vspace{-5mm}
\end{table}

\begin{table}
\begin{center}
\begin{tabular}{l | c | c | c}
\hline
Category & Baseline & Ours & Delta \\
\hline
Artwork &	42.9\% &	\textbf{75.5\%} &	76.0\% \\
Beds \& Bed Frames &	16.7\% &	\textbf{45.7\%} &	173.7\% \\
Benches &	14.7\% &	\textbf{51.4\%} &	249.7\% \\
Cabinets \& Storage &	22.9\% &	\textbf{64.6\%} &	182.1\% \\
Candles &	\textbf{60.0\%} &	40.0\% &	-33.3\% \\
Chairs &	22.4\% &	\textbf{63.3\%} &	182.6\% \\
Curtains \& Drapes &	39.6\% &	\textbf{89.8\%} &	126.8\% \\
Dressers &	91.7\% &	\textbf{100.0\%} &	9.1\% \\
Fireplaces &	40.4\% &	\textbf{72.9\%} &	80.4\% \\
Folding Chairs \& Stools &	37.9\% &	\textbf{46.7\%} &	23.2\% \\
Lighting &	33.3\% &	\textbf{69.4\%} &	108.4\% \\
Mirrors &	30.6\% &	\textbf{49.0\%} &	60.1\% \\
Ottomans &	50.0\% &	\textbf{80.0\%} &	60.0\% \\
Pillows &	57.4\% &	\textbf{75.5\%} &	31.5\% \\
Rugs &	71.4\% &	\textbf{85.7\%} &	20.0\% \\
Shelving &	12.5\% &	\textbf{42.9\%} &	243.2\% \\
Sofas &	18.4\% &	\textbf{38.8\%} &	110.9\% \\
Table \& Bar Stools &	40.9\% &	\textbf{82.2\%} &	101.0\% \\
Tables &	16.3\% &	\textbf{40.8\%} &	150.3\% \\
Vases &	48.9\% &	\textbf{69.4\%} &	41.9\% \\
\hline
Overall &	37.3\% &	\textbf{64.2\%} &	72.1\% \\
\hline
\end{tabular}
\end{center}
\caption{Human Judgements for Shop-the-Look measuring Precision@5 comparing unified embedding vs existing specialized embedding. We see that our new unified embedding significantly improves the relevance overall with wins in all categories except one.}
\label{tab:stl_eval}
\vspace{-5mm}
\end{table}

\subsection{A/B Experiments}
A/B experiments at Pinterest are the most important criteria for deploying changes to production systems.

\begin{figure}[t]
\begin{center}
\includegraphics[width=1.0\linewidth]{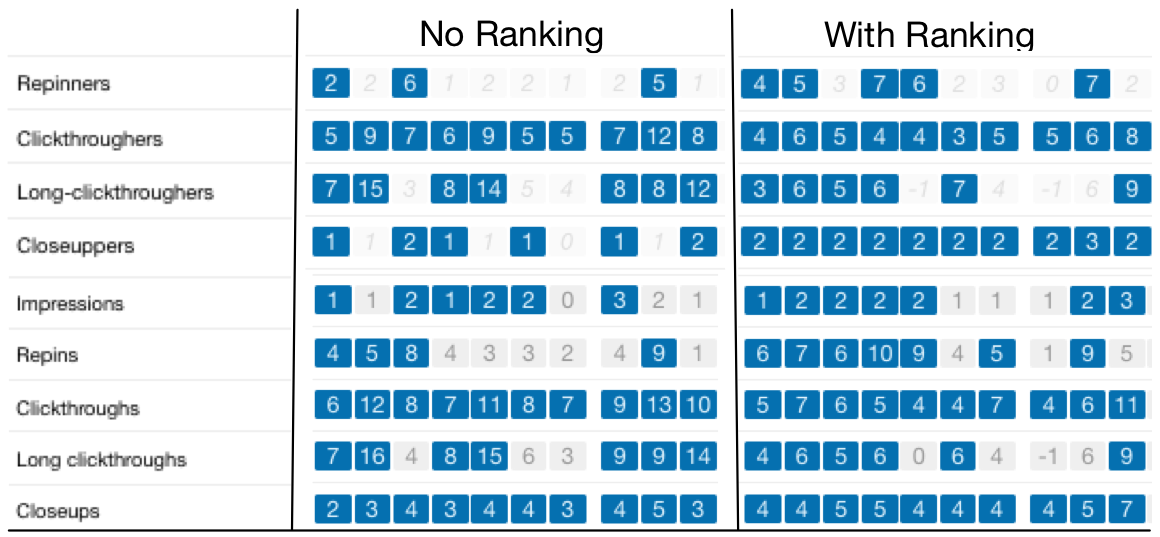}
\end{center}
   \caption{A/B experiment results on Flashlight showing changes in metrics (Blue highlights statistically significant changes) for users across days in the experiment (to diagnose novelty effects if any). We see significant lifts in engagement propensity and volume with our unified embedding compared to the existing specialized embedding.}
\label{fig:flashlight_ab}
\end{figure}

Flashlight A/B experiment results of our unified embedding vs the old specialized embedding are shown in Figure~\ref{fig:flashlight_ab}. We present results on two treatment groups: (1) A/B experiment results on Flashlight with ranking disabled and (2) A/B experiment results on Flashlight with ranking enabled. Flashlight candidate generation is solely dependent on the embedding and as such, when disabling ranking we can see impact of our unified embedding without the dilution of the end-to-end system. For deployment, we look at the treatment group with ranking enabled. In both cases, our unified embedding significantly improves upon the existing embedding. We see improvement in top-line volume metrics of impressions, closeups, repins (action of saving a Pin to a Board), clickthroughs, and long clickthoughs (when users remain off-site for an extended period of time~\cite{relatedpins}) along with improvement in top-line propensity metrics (percent of Flashlight users who do a specific action daily) of closeuppers, repinners, clickthroughers, and long clickthroughers.

Lens A/B experiment results of our unified embedding vs the old specialized embedding are shown in Table~\ref{tab:lens_ab}. As a newer product, Lens A/B experiment results are generated via a custom analysis script hence the difference in reporting between Flashlight and Lens. Similar to the Flashlight A/B results, we see significant improvement to both our engagement and volume metrics when replacing the existing embedding with our unified embedding for Lens.

Shop-the-Look had not launched to users when experimenting with our unified embedding and as such, no A/B experiments could be run. As a prototype, we focused on relevance and as such the human judgement results were used as the launch criteria.

\begin{figure}[t]
\begin{center}
\includegraphics[width=1.0\linewidth]{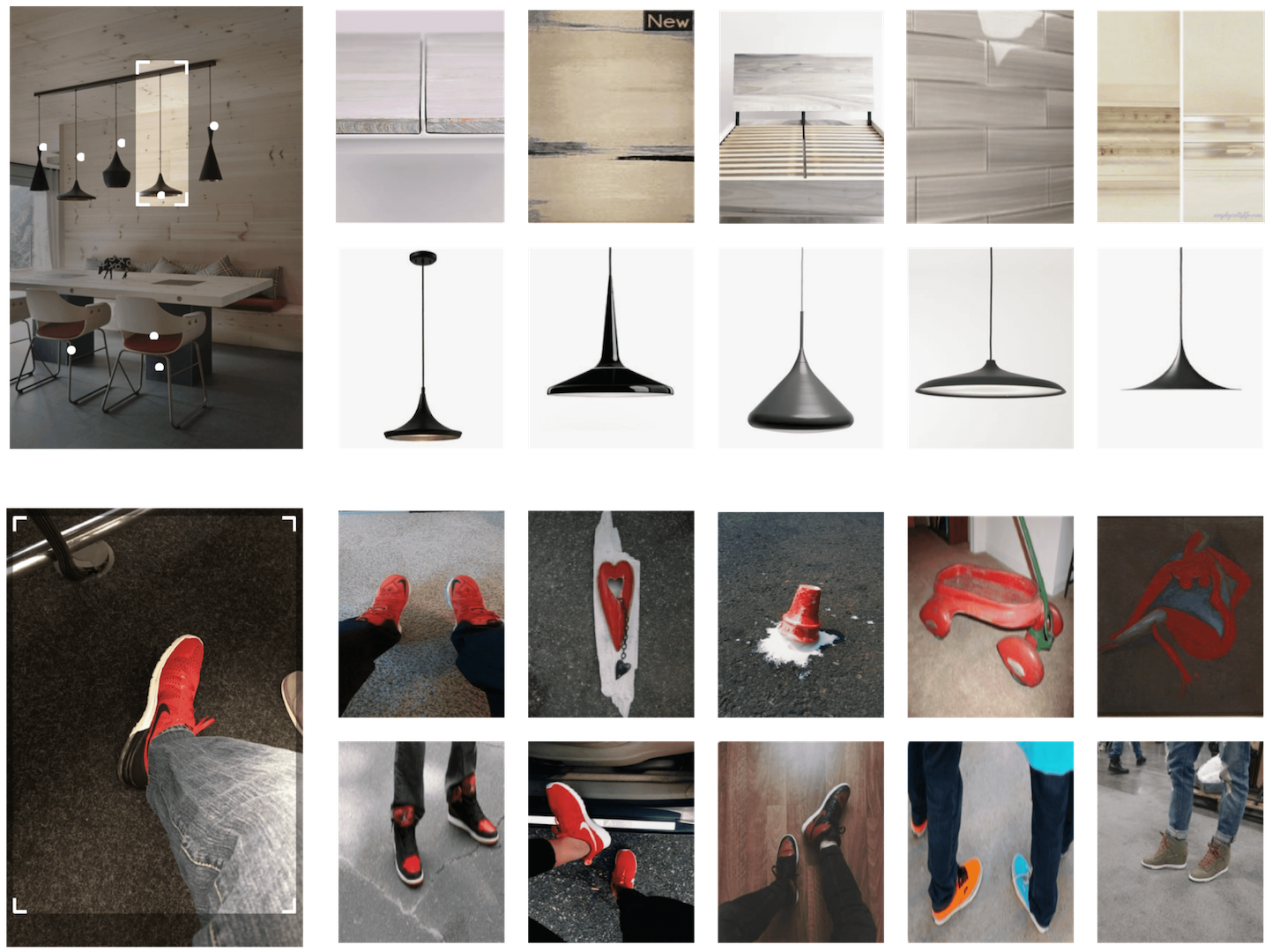}
\end{center}
   \caption{Qualitative results comparing the old embeddings vs our new multi-task embeddings in Flashlight. For each query on the left, the results from old embeddings are shown on the top row, and the new embeddings are shown on the bottom row.}
\label{fig:productionized}
\end{figure}

Given the significantly positive relevance human judgements and A/B experiments results, we deployed our unified embedding to all the visual search products, replacing the specialized embedding with one representation that outperformed on all tasks. Qualitative results of our unified embedding compared to the old specialized embeddings can be seen in Figure~\ref{fig:productionized}.

\begin{table}
\begin{center}
\begin{tabular}{c c c c c c}
\hline
Closeuppers & Repinner & Clickthrougher   \\
\hline
+16.3\% & +26.7\% & +24.3\% \\
\hline
\hline
Closeup & Repin & Clickthrough \\
\hline
+32.7\% & +46.7\% & +35.0\% \\
\hline
\end{tabular}
\end{center}
   \caption{A/B experiment results on Lens. We see significant lifts in engagement propensity and volume with our unified embedding compared to the existing specialized embedding.}
\label{tab:lens_ab}
\vspace{-10mm}
\end{table}

%% file: main_conclusion.tex
\section{Conclusion}
Improving and maintaining different visual embeddings for multiple customers is a challenge. At Pinterest, we took one step to simplifying this process by proposing a multi-task metric learning architecture capable of jointly optimizing multiple similarity metrics, such as browsing and searching relevance, within a single unified embedding. To measure the efficacy of the approach, we experimented on three visual search systems at Pinterest, each with its own product usage. The resulting unified embedding outperformed all specialized embedding trained with individual task in comprehensive evaluations, such as offline metrics, human judgements and A/B experiments. The unified embeddings are deployed at Pinterest after observing substantial improvement in recommendation performance reflected by better user engagement across all three visual search products. Now with only one embedding to maintain and iterate, we have been able to substantially reduce experimentation, storage, and serving costs as our visual search products rely on a unified retrieval system. These benefits enable us to move faster towards our most important objective -- to build and improve products for our users.